\documentclass{ecai} 
\usepackage{setspace}

\usepackage{latexsym}
\usepackage{amssymb}
\usepackage{amsmath}
\usepackage{amsthm}
\usepackage{booktabs}
\usepackage{enumitem}
\usepackage{graphicx}
\usepackage{color}
\usepackage{xcolor}
\usepackage{tabularx}
\usepackage{amssymb}
\usepackage{arydshln}
\usepackage{multirow}

\usepackage{tabu}                     
\usepackage{multirow}                 
\usepackage{multicol}                 
\usepackage{multirow}                
\usepackage{float}                    
\usepackage{makecell}                 
\usepackage{booktabs}                 
\usepackage{fancyhdr}
\usepackage{graphicx}
\usepackage{amsmath}
\usepackage{titlesec}

\newcommand{\BibTeX}{B\kern-.05em{\sc i\kern-.025em b}\kern-.08em\TeX}

\begin{document}


\begin{frontmatter}


\paperid{123} 


\title{Enhancing Text-to-SQL Capabilities of Large Language Models via Domain Database Knowledge Injection}



\author[A]{\fnms{Xingyu}~\snm{Ma}\footnote{Equal contribution.}}
\author[B]{\fnms{Xin}~\snm{Tian}\footnotemark}
\author[C]{\fnms{Lingxiang}~\snm{Wu}} 
\author[C]{\fnms{Xuepeng}~\snm{Wang}\thanks{Corresponding Author. Email: xuepeng.wang@ia.ac.cn.}}
\author[A]{\fnms{Xueming}~\snm{Tang}\thanks{Corresponding Author. Email: xmtang@hust.edu.cn.}}
\author[C,D,E]{\fnms{Jinqiao}~\snm{Wang}}
\address[A]{School of Cyber Science Engineering, Huazhong University of Science and Technology}
\address[B]{Wuhan AI Research}
\address[C]{Foundation Model Research Center, Institute of Automation, Chinese Academy of Sciences}
\address[D]{School of Artificial Intelligence, University of Chinese Academy of Sciences}
\address[E]{Peng Cheng Laboratory}


\begin{abstract}
Text-to-SQL is a subtask in semantic parsing that has seen rapid progress with the evolution of Large Language Models (LLMs). However, LLMs face challenges due to hallucination issues and a lack of domain-specific database knowledge(such as table schema and cell values). As a result, they can make errors in generating table names, columns, and matching values to the correct columns in SQL statements. This paper introduces a method of knowledge injection to enhance LLMs' ability to understand schema contents by incorporating prior knowledge. This approach improves their performance in Text-to-SQL tasks. Experimental results show that pre-training LLMs on domain-specific database knowledge and fine-tuning them on downstream Text-to-SQL tasks significantly improves the \textsc{Execution Match (EX)} and \textsc{Exact Match (EM)} metrics across various models. This effectively reduces errors in generating column names and matching values to the columns. Furthermore, the knowledge-injected models can be applied to many downstream Text-to-SQL tasks, demonstrating the generalizability of the approach presented in this paper.
\end{abstract}

\end{frontmatter}


\section{Introduction}

With the advancement of LLMs, certain large models like WizardCoder\citep{luo2023wizardcoder} and CodeLlama\citep{roziere2023code} have achieved satisfactory results in Text-to-SQL tasks. However, applying LLMs to Text-to-SQL tasks presents challenges due to polysemous terms and insufficient semantic information in column and table names.
\begin{itemize}
    \item \textbf{Example of polysemous column names:} A column named \texttt{"state"} can have several meanings (e.g. condition, situation, political entity and so on); If \texttt{"New York State"} is given, LLMs can easily understand the meaning of \texttt{"state"}; 
    \item \textbf{Example of columns with insufficient semantic clarity:} A column named \texttt{"b\_fund\_stock\_bond\_share\_d"} stands for \texttt{"daily total fund stock and bond position ratios"}. This is a column name from a real-world database, and it's difficult to determine the meaning of the column based on the few words in the column name alone. However, if some values of the column are provided, then the model can easily understand the meaning of the column. 
\end{itemize}
This could lead to the model being prone to two types of errors.: 1) Errors in generating column names and table names, 2) Mismatches between cell values, column names, and table names, as shown in Table \ref{tab1}.
\begin{table}[htbp]
\label{tab1}
\caption{The following examples, highlighted in green, illustrate common errors in Text-to-SQL Tasks: \textbf{Case 1} demonstrates the incorrect generation of column names, while \textbf{Case 2} demonstrates a mismatch between column names and cell values.
}
\vspace{10pt}
\centering
\resizebox{1\columnwidth}{!}{\begin{tabular}{l}
\toprule
\multicolumn{1}{c}{\textbf{Case 1: Columns Generation Error}}\\
\midrule
\textcolor{red}{\emph{\textbf{List the arrival date and the departure date for all the dogs}}}\\
\colorbox{yellow}{\textbf{Gold:}} \texttt{SELECT \colorbox{green}{date\_arrived},  \colorbox{green}{date\_departed} FROM Dogs}\\
\colorbox{yellow}{\textbf{Predict:}} \texttt{SELECT \colorbox{green}{date\_of\_arrival},  \colorbox{green}{date\_of\_departure} FROM Dogs}\\
\toprule
\multicolumn{1}{c}{\textbf{Case 2: Mismatch between Column Name and Cell Value}}\\
\midrule
\textcolor{red}{\emph{\textbf{What country is Jetblue Airways affiliated with?}}}\\
\colorbox{yellow}{\textbf{Gold:}} \texttt{SELECT Country FROM AIRLINES WHERE \colorbox{green}{Airline}  =  "JetBlue Airways"}\\
\colorbox{yellow}{\textbf{Predict:}}\texttt{SELECT Country FROM airlines WHERE \colorbox{green}{Abbreviation}  =  "Jetblue Airways"}\\
\bottomrule
\end{tabular}}
\end{table}

Numerous studies have sought to resolve these two issues by regulating the decoding rules\citep{scholak2021picard}, achieving some progress. However, such optimizations of decoding rules have not enhanced the base model's comprehension of databases. While they can rectify certain errors in generating column names and table names, matching cell values with columns remains challenging to address through restricted decoding rules. Further, there are efforts to optimize this problem through schema linking\citep{wang2019rat}, but the cell values and column names extracted via this method do not consistently reach a high level of accuracy. In LLMs, due to the pretrained nature of the models, it is difficult to alter the model structure to implement schema linking, with the only option being to incorporate schema information into prompts. However, LLMs lack the domain-specific database knowledge, which impairs their ability to comprehend and apply schema information provided within prompts. Therefore, we aim to enhance the ability of LLMs to comprehend database schema and cell values through the injection of domain database knowledge to optimize this issue. This approach is intended not only to improve the LLMs' comprehension of databases but also to achieve a certain level of optimization for such challenges. 

Although some work focuses on enhancing the understanding of data tables through methods like table pretraining\citep{yin2020tabert,yu2020grappa,deng2020structure,shi2021learning}, these efforts are general, based on extensive datasets, and lack specificity; they fail to solve domain-specific issues like incorrect generation of column and table names. Consequently, this paper seeks to explore how to specifically inject domain database schema and cell values as prior knowledge into LLMs, thereby improving their capacity to apprehend domain database schema and cell values accurately, enabling the correct identification and generation of table names, column names, and cell values. Our primary thoughts and work are as follows:
\begin{itemize}
    \item \textbf{Semantic knowledge fusion.} In dealing with polysemy and phrases or vocabulary with ambiguous semantics, semantic fusion is a commonly used solution\citep{lyu2021let}. This paper draws on this idea, hoping to enhance the semantic information of column names and table names by modeling the relationships between column names, table names, and cell values.
    \item \textbf{Improving the co-occurrence frequency between table name and column names.} When LLMs predict the next token, they tend to output tokens that have a higher co-occurrence frequency with the previous tokens\citep{kang2023impact}. Therefore, by modeling the relationships between column names and table names, we aim to increase their co-occurrence frequency, thus making the model more inclined to generate the correct column and table names when deriving table names from column names or vice versa.
\end{itemize}
This paper conducts experiments on several open-source LLMs using the constructed dataset. The experimental results indicate that the methods presented in this paper can improve the EX and EM metrics, and reduce the instances of errors in the generation and matching of column names, table names, and cell values.


\section{Methodology}
We will introduce our method through three subsections. In Section 2.1, we will present which features of the database schema and cell values can be utilized to build prior knowledge. Section 2.2 will detail how to leverage these features to construct prior knowledge, while Section 2.3 will outline the specific design methods for each task.
\subsection{What Knowledge can LLMs Learn From a Database?}
Firstly, let's define the meanings of two types of knowledge:
\begin{itemize}
    \item \textbf{Table and column names semantic knowledge:} Database administrators follow certain conventions when creating tables, often using column and table names that contain semantic information. For example, a table named "user\_info" can be inferred to be a user data table, this is the semantic knowledge of column and table names. However, there still exists a significant number of tables and columns where semantic information is incomplete, and we expect the model to utilize cell values to enhance the semantic knowledge of such column and table names.
    \item \textbf{Schema knowledge:} The schema of a database refers to information such as table names, column names, and types of columns. We refer to this type of knowledge as schema knowledge.
\end{itemize}

\begin{enumerate}

\item{\textbf{LLMs can learn the semantic knowledge of tables and columns using cell values}}
\begin{figure}[htbp]
    \centering
    \includegraphics[width=8.3cm]{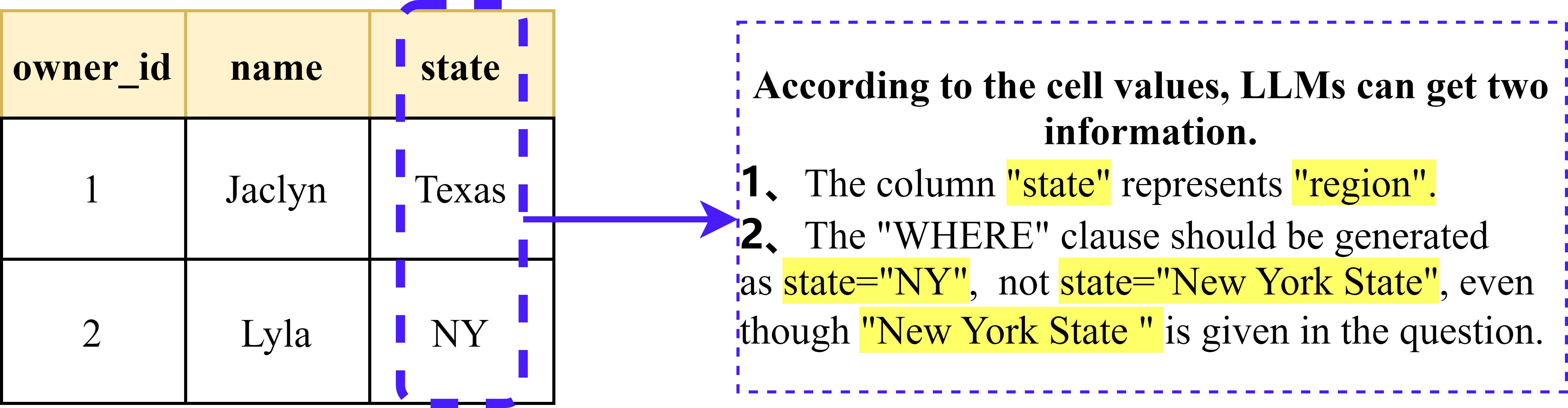}
    \caption{Learning Column Names Semantic Knowledge}
    \vspace{10pt}
\end{figure}

The data type of a column in a database table is typically a collection of data of the same type, so we can use the cell values contained in that column to enhance the semantic information of the column name. For columns with polysemous words or columns with incomplete semantic information (especially in practical scenarios where column names in data tables do not have clear semantic information, such as, the column "\texttt{b\_fund\_stock\_bond\_industry\_share\_d}" describes "\texttt{ratios of daily total fund stock and bond position}", which LLMs cannot intuitively deduces), this can effectively enhance the model's understanding of the column. This, in turn, facilitates the large model's ability to understand prompts and to extract and match cell values and columns from questions. This idea is similar to some methods that use semantic fusion to solve problems with polysemy\citep{lyu2021let}. For example, when a prompt or question mentions the \texttt{"state"} column, since \texttt{"state"} is a polysemous word, LLMs do not know what \texttt{"state"} means or what cell values \texttt{"state"} might contain. However, if a few value informations are given, such as \texttt{"Texas"} and \texttt{"NY"}, LLMs can learn that the \texttt{"state"} column denotes a location. At the same time, LLMs can also learn that in this database, the value for \texttt{"state"} is \texttt{"NY"} rather than \texttt{"New York state"}. Thus, when generating cell values, the occurrence of generating \texttt{"New York state"} is reduced, as shown in Figure 1.

\begin{figure}[htbp]
    \centering
    \includegraphics[width=8.3cm]{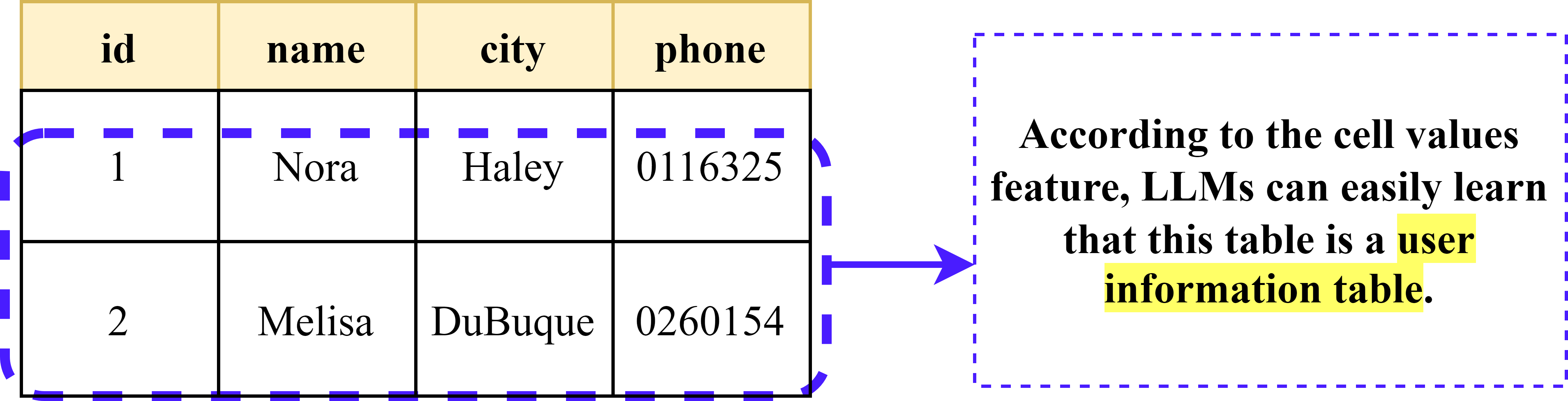}
    \caption{Learning Table Names Semantic Knowledge}
    \vspace{10pt}
    \label{fig:galaxy}
\end{figure}

Similarly, each row of data in a table represents the attributes of an item. Therefore, it is also easy to determine the type of the table based on the cell values it contains, which enhances the model's ability to understand the table. This is likely to have a certain effect on schema linking, as illustrated in Figure 2.

\item {\textbf{LLMs can learn schema knowledge to improve the co-occurrence frequency between table names and column names}}
\begin{figure}[htbp]
    \centering
    \includegraphics[width=8.5cm]{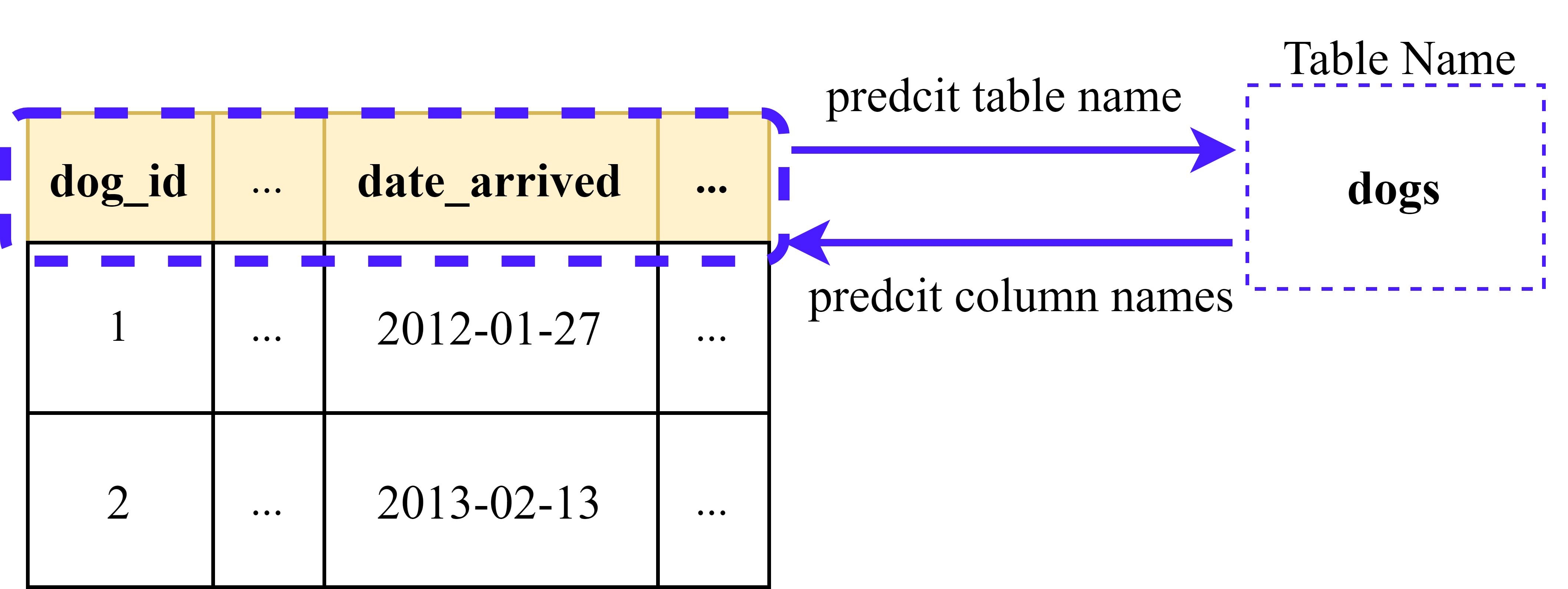}
    \caption{Learning Schema Knowledge}
    \vspace{10pt}
    \label{fig:galaxy}
\end{figure}

We understand that language models primarily learn from the co-occurrence relationships of tokens and tend to generate vocabulary with higher co-occurrence frequency when predicting the next token\citep{kang2023impact}. Consequently, we aim to increase the co-occurrence frequency between column names and table names to make the model more apt to generate the correct names when predicting them. As depicted in Figure 3, we aim to boost the number of co-occurrences between column names and table names by constructing a mutual predictive relationship between them. Table 1 illustrates a scenario where the question refers to \texttt{"the arrival date"}. but the LLMs generate \texttt{"date\_of\_arrival"} instead of the actual column name, which is \texttt{"date\_arrived"}. This discrepancy arises because, although the LLMs understand the question, they are not acquainted with the database's schema information, resulting in \begin{figure*}[htbp]
    \centering
    \includegraphics[width=18cm]{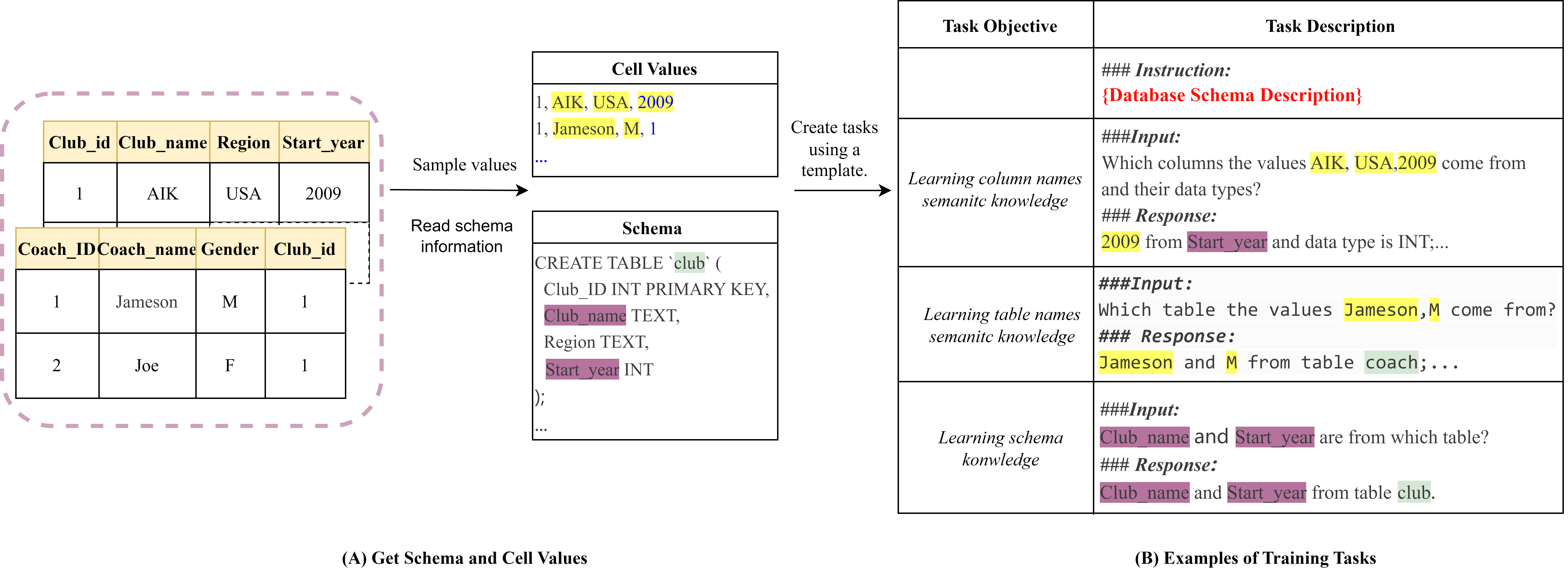}
    \caption{Overview of method, The cell values data from the database tables are highlighted with a yellow background, the column names data are highlighted with a purple background, and the table names data are highlighted with a light green background. Our task data format construction refers to the data structure format of the Alpaca\citep{wang2022self} directive.}
    \vspace{10pt}
    \label{fig:galaxy}
\end{figure*} a  mismatch when generating column names. If the co-occurrence frequency between \texttt{"date\_arrived"} and \texttt{"Dogs"} were enhanced, the model would be more inclined to generate the column name as \texttt{"date\_arrived"}.
\end{enumerate}

\subsection{How to Build Training Tasks to Inject Domain Knowledge?}
From the time the T5 model organized natural language processing (NLP) tasks as end-to-end text generation tasks\citep{raffel2020exploring}, to the present day with the emergence of various LLMs, NLP tasks have almost all become end-to-end text generation tasks. Therefore, although the idea of training tasks is derived from classic NLP tasks such as Named Entity Recognition (NER), the form of training is also conducted as end-to-end text generation tasks. As shown in Figure 4, we first obtain the schema information and cell values from the database, then use pre-constructed question-and-answer templates to fill in the cell values and schema information into the training templates to build training data. Our template construction mainly focuses on three task objectives: enhancing column name semantics, enhancing table name semantics, and increasing the co-occurrence frequency between column names and table names through learning schema knowledge.

Specifically, given a database $\mathcal D_{i}\in\mathbb{D}$, the schema information of the database $\mathcal D_{i}$ is $\mathcal S_{i}\in\mathbb{S}$, where $|\mathbb{D}|=|\mathbb{S}|$, and $i \in \left[1, |\mathbb{D}|\right]$, The database $\mathcal D_{i}$ contains $K$ tables $T_{j}$, where $j \in \left[1, K\right]$, Each table has value information $\mathcal V_{i,j} \in \mathbb{V}_{i}$, We have a set of question templates $\mathbb T_{q}=\{t_{q,1},t_{q,2}...t_{q,N}\}$ and and a set of answer templates $\mathbb T_{a}=\{t_{a,1},t_{a,2}...t_{a,N}\}$, from which we construct the following training objectives.

\begin{eqnarray}
\mathcal{G}_m=\delta (t_{a,m},\mathcal{S}_i,\mathcal{V}_{i,j})\\
\widetilde{\mathcal{P}_m}=\mathcal{M}(\sigma(t_{q,m},\mathcal{S}_i,\mathcal{V}_{i,j}))\\
\min_{\mathcal{M}}\sum_{m=1}^N\mathcal{L}(\mathcal{G}_m,\widetilde{\mathcal{P}_m})
\end{eqnarray}

Where $\delta(\cdot,\cdot,\cdot)$ denotes the construction of the answer $\mathcal{G}_m$ by utilizing the answer template $t_{a,m}$, along with the schema $\mathcal{S}_i$ and cell vaules $\mathcal{V}_{i,j}, \sigma(\cdot,\cdot,\cdot)$ represents the construction of a question using the question template $t_{q,m}$, schema $\mathcal{S}_i$ and cell values $\mathcal{V}_{i,j}$, $\mathcal{M}$ signifies the base model. $\mathcal{L}$ denotes the loss function, $\widetilde{\mathcal{P}_m}$ represents the model's predicted results, and our training objective is to minimize the error between the model's predicted results $\widetilde{\mathcal{P}_m}$ and the actual results $\mathcal{G}_m$.

\subsection{Construction of Tasks and Cell Values Sampling}
As shown in Figure 4(B), we aim to enhance the model's Text-to-SQL capabilities through three learning objectives. For each learning objective, we have designed some learning tasks based on classical NLP task concepts, such as ideas from NER tasks, text clustering, etc. The specific task designs are as follows.

\textbf{Objective 1.  \emph{Learning column names semantic knowledge}}

The purpose of this task is to utilize cell values to enhance the semantic information of column names, thereby improving the model's understanding of domain-specific database column names. This is intended to reduce the incidence of incorrect column matching with cell values, such as \texttt{Abbreviation="Jetblue Airways"} showing in Table 1. The specific task design is as follows:
\begin{itemize}
\item Sample cell values $\mathcal{V}_i$ from a column and train the model to generate the corresponding column name $\mathcal{C}_i$ for those cell values. Formulation as $\mathcal{\max}\mathbb{P}(\mathcal{C}_i|\mathcal{V}_i=\{v_{1}, v_{2}...v_{n}\})$.
\item Sample cell values $\mathbb{V}={\mathcal{\{V\}}}_{1}^{n}$ from $n$ columns and train the model to cluster these values according to their column names ${\mathcal{\{C\}}}_{1}^{n}$. Formulation as $\mathcal{V}_{1},...,\mathcal{V}_{n}=\zeta^{\star}(shuffle(\mathbb{V}))$, where $\mathcal{V}_{n}=\{v_{1},...v_{\|\mathcal{V}_n\|}\}$, $\zeta^\star$ is base model and the $shuffle$ is shuffle function to mix the values.
\item Sample cell values $\mathcal{V}_i$ and column name $\mathcal{C}_i$ and train the model to determine whether the cell values belong to that column.
Formulation as follows, Where $\mathbb{I}$ is indicator function and $\mathcal{C}_j$ is the true column name of $\mathcal{V}_i$.
\begin{eqnarray}
\max\mathbb{P}(\mathbb{I}(\mathcal{C}_i,  \mathcal{C}_j)|\mathcal{V}_{i}, \mathcal{C}_i)\\
s.t. \quad\left\{\begin{matrix}
 \mathbb{I}(\mathcal{C}_i,  \mathcal{C}_j)=1, & if\quad \mathcal{C}_i=\mathcal{C}_j\\ 
 \mathbb{I}(\mathcal{C}_i,  \mathcal{C}_j)=0, & if\quad \mathcal{C}_i\neq\mathcal{C}_j
\end{matrix}\right.
\end{eqnarray}
\item Given cell values $\mathcal{V}_i$ and column name $\mathcal{C}_i$, train the model to predict the data type of that column.
Formulation as $\max\mathbb{P}(\mathcal{T}_i|\mathcal{C}_i,\mathcal{V}_i)$, where $\mathcal{T}_i$ is the data type, such as \texttt{INT}.
\end{itemize}

\textbf{Objective 2. \emph{Learning table names semantic knowledge}} 

The purpose of this task is to enhance the model's understanding of tables. For instance, if there is a clear distinction between \texttt{user} tables and \texttt{product} tables, the model can benefit by preliminarily determining which table a query should reference based on the values mentioned in the user's question. The specific task design is as follows:
\begin{itemize}
\item Sample a row of data $\mathcal{R}_i=\{v_{i,1},v_{i,2}...v_{i,n}\}$ from a table and train the model to generate the corresponding table name $\{T_{j}\}_{1}^{\|\mathcal{D}\|}$. where $\|\mathcal{D}\|$ is the number of tables,Formulation as $\max\mathbb{P}(T_j|\mathcal{R}_i)$.
\item Sample some cell values from each table and train the model to cluster the cell values that belong to the same table together. Formulation as follows:
\begin{eqnarray}
\mathcal{V}_1,..,\mathcal{V}_{\|\mathcal{D}\|} = \mathbb{\zeta^\star}(shuffle(\mathbb{V})) \\
\mathbb{V} = \{\mathcal{V}_1,\mathcal{V}_2,..., \mathcal{V}_{\|\mathcal{D}\|}\}, \quad 
 \mathcal{V}_{\|\mathcal{D}\|} \in T_{\|\mathcal{D}\|} \\
\mathbb{T} = \{T_1,T_2,...,T_{\|\mathcal{D}\|}\}
\end{eqnarray}
\end{itemize}
Where $\|\mathcal{D}\|$ is the number of tables, $\mathbb{T}$ is a set of tables , $\mathbb{V}$ is the set of values from the $\mathbb{T}$, $shuffle$ is a function used to mix $\mathbb{V}$, and $\zeta^\star$ is base model.

\textbf{Objective 3. \emph{Learning schema knowledge}} 

The goal of this task is to increase the frequency of co-occurrence between table names and column names by learning schema knowledge, so that when a table name is mentioned in the previous context, the model is more inclined to generate the correct column name for the next token prediction; likewise, when a column name is generated in the previous context, it is more inclined to predict the correct table name for the next token. The specific forms of the task are as follows.
\begin{itemize}
\item Sample column names $\mathcal{C}_{i}$ from a table $T_{i}$ and train the model to generate the table name for these columns. Formulation as $\mathcal{\max}\mathbb{P}(T_{i}|\mathcal{C}_i\})$, where $\mathcal{C}_i \in \{c_{i,1},c_{i,2},...,c_{i,\|T_{i}\|}\}$.
\item Sample a few tables $\mathbb{T}$ from the database and then sample some column names $\mathbb{C}_{i}$ from these tables, shuffle these column names together, and train the model to cluster the columns according to their table names so they are grouped with the same table name. Formulation as $C_{1},..,C_{\|\mathbb{T}\|}=\zeta^{\star}(shuffle(\mathbb{C}_i)\})$, where $shuffle$ is a function used to mix $\mathbb{C}$, $\zeta^{\star}$ is base model and $\mathcal{C}_{\|\mathbb{T}\|} \in T_{\|\mathbb{T}\|}$ 
\item Randomly sample two tables and determine whether they can perform a \texttt{JOIN} operation to enhance understanding of multi-table query tasks. Formulation as follows:
\begin{eqnarray}
\max\mathbb{P}(\mathbb{I}(T_i, T_j)|T_i, T_j)\\
s.t. \quad\left\{\begin{matrix}
 \mathbb{I}(T_i, T_j)=1, & if\quad T_{i}^{fk}=T_{j}^{fk}\\ 
 \mathbb{I}(T_i, T_j)=0, & if\quad T_{i}^{fk}\neq T_{j}^{fk}
\end{matrix}\right.
\end{eqnarray}
\end{itemize}
Where $\mathbb{I}$ is an indicator function, $T$ is a table, and $fk$ is foreign key of the table.

In order to increase the co-occurrence of column names and table names, and to align with the style of SQL, we set the answers to be similar to SQL statements, for example, "\texttt{NAME FROM USER}".

\textbf{Cell values sampling.} Due to the large number of rows in a table, we cannot utilize all rows. Therefore, we initially sample $\mathcal{N}$ rows of data to serve as a subtable of the table. Specifically, all rows of the table are clustered into $\mathcal{K}$ categories, and then $\mathcal{L}$ rows of data are selected from each cluster to construct the subtables. When selecting a row of data, we opt for the first $\mathcal{L}$ rows with the least amount of missing values, as such data contains more information. Thus, $\mathcal{N}=\mathcal{K*L}$. Furthermore, since many cell values are numerical and do not contain semantic information that can affect the training of the model, we control the proportion of numbers when sampling cell values so that it does not exceed 50\%.

\section{Experiments}
\subsection{Datasets}
The experimental dataset in this paper utilizes the \textsc{Spider} dataset \citep{yu2018spider}, which is a large-scale, complex, multi-domain dataset and also the most widely used in the Text-to-SQL task. The \textsc{Spider} dataset consists of training, development, and test sets, with no overlap between the training set, development set, and test set. This paper uses the databases included in the \textsc{Spider} training and development sets to construct training data according to the knowledge injection fine-tuning task introduced in Section 2. A total of 49,878 task data entries were constructed, including 17,954 entries related to the co-occurrence frequency enhancement task for column and table names, and 31,924 entries related to the semantic enhancement task for column and table names. Then, using 8,926 entries from the \textsc{Spider} training set for the second-phase data to fine-tune the downstream Text-to-SQL task, and the \textsc{Spider} development and test sets for effectiveness validation.

\subsection{Evaluation Metrics}
We use two of the most mainstream evaluation metrics, \textsc{Execution Match (EX)} and \textsc{Exact Match (EM)}, to assess the experimental results. The EM metric represents the proportion of generated SQL statements that exactly match the labeled SQL statements. The EX metric involves comparing the execution results of the generated SQL statements with those of the labeled SQL statements after execution; if the results are the same, it is deemed correct.
\subsection{Models, Baseline and Environments}
Our experiments primarily utilized open-source code models deepseek-coder-6.7b\citep{guo2024deepseek}, CodeLlama-7b-Instruct, CodeLlama-13b-Instruct\citep{roziere2023code}, and WizardCoder-15B-V1.0\citep{luo2023wizardcoder}. To validate the effectiveness of our method on general models, we also experimented with the general model Baichuan2-7B-Chat\citep{yang2023baichuan}. The baseline code for our downstream Text-to-SQL tasks was DB-GPT-HUB\footnote{\url{https://github.com/eosphoros-ai/DB-GPT-Hub}}, an open-source code aimed at Text-to-SQL tasks based on LLMs. The fine-tuning code for phase one database understanding enhancement tasks was implemented by us. Due to time constraints, all experimental code parameters were kept the same, with adjustments made only to batch size and deepspeed parameters in response to Out-Of-Memory issues, indicating that our experimental parameter settings might not be optimal. Our experimental environment was based on the transformers and Pytorch frameworks, with NVIDIA A800-SXM4-80GB GPUs, and each model was trained for 15 epochs.
\subsection{Results}

\begin{table}[]
\centering
\caption{The main experimental results, where \textsc{Spider-dev} denotes the development set of \textsc{SPIDER}, and \textsc{Spider-Syn} is a dataset created by substituting synonyms in the questions of the spider development set. We will introduce this dataset in the robustness testing section. The model names in the table have been abbreviated for brevity. "\emph{KE}" stands for "Knowledge Enhanced," indicating the evaluation results of models that have been augmented with domain-specific database knowledge. Results not marked with "\emph{KE}" represent the evaluation outcomes of the baseline.}
\vspace{10pt}
\renewcommand{\arraystretch}{1.2}
\begin{tabular}{lcccc}
\toprule
\multirow{2}{*}{\textbf{Models}}                            & \multicolumn{2}{c}{\textbf{\textsc{Spider-dev}}}                             & \multicolumn{2}{c}{\textbf{\textsc{Spider-Syn}}} \\
\cline{2-5}
\rule{0pt}{12pt}
                                                            & \textbf{EX\%}                    & \textbf{EM\%}                    & \textbf{EX\%}        & \textbf{EM\%}        \\
\midrule
T5-3B\citep{scholak2021picard}             & 74.4                           & 71.5                           & -                  & -                  \\
T5-3B$_\emph{KE(ours)}$                 & \textbf{75.6}                 & \textbf{72.1} & -                  & -                  \\
T5-3B$_{PICARD}$\citep{scholak2021picard} & 79.3                           & 75.5                           & -                  & -                  \\
T5-3B$_{PICARD+\emph{KE(ours)}}$          & \textbf{80.7} & \textbf{75.5} & -                  & -                  \\
\hdashline
deepseek-6.7b                                                & 77.8                           & 73.9                           & 64.4                  & 58.6                  \\
deepseek-6.7b$_{+\emph{KE(ours)}}$               & \textbf{80.9} & \textbf{73.6} & \textbf{65.6}                  & \textbf{58.7} \\

CodeLlama-7b                                                & 74.5                           & 70.2                           & 59.2                  & 53.5                  \\
CodeLlama-7b$_{+\emph{KE(ours)}}$               & \textbf{76.6} & \textbf{73.0} & \textbf{62.1}                  & \textbf{55.8}                  \\
CodeLlama-13b                                               & 75.7                           & 71.1                           & 61.6                  & 55.6                  \\
CodeLlama-13b$_{+\emph{KE(ours)}}$              & \textbf{78.3} & \textbf{73.5} & \textbf{65.4}                  & \textbf{60.5}                  \\
WizardCoder-15B                                             & 75.4                           & 70.5                           & 58.7                  & 52.6                  \\
WizardCoder-15B$_{+\emph{KE(ours)}}$          & \textbf{76.6} & \textbf{72.9} & \textbf{60.8}                  & \textbf{54.2}                  \\
Baichuan2-7B                                                & 68.2                           & 64.0                           & 53.6                  & 49.1                  \\
Baichuan2-7B$_{+\emph{KE(ours)}}$               & \textbf{71.3} & \textbf{66.5} & \textbf{54.6}                 & \textbf{50.1}   \\
\bottomrule
\end{tabular}
\label{tab:accents}
\end{table}

\textbf{Main Results.}
The experimental results are presented in Table 2. Initially, it is observable that, on the T5-3B model, the method proposed in this study demonstrates an improvement of 1.2\% in the \textsc{EX} metric and 0.6\% in the \textsc{EM} metric compared to the baseline model. After incorporating PICARD to enforce decoding rules, our approach still yields enhancements in both the EX and EM metrics.

Moreover, experiments on open-source LLMs have shown significant improvements over the baseline models in terms of the Exact Match (EX) and Execution Match (EM) metrics. Firstly, the deepseek-coder-6.7b model saw the largest increase in the EX metric, with an improvement of 3.1\%; on the CodeLlama-13b model, the EM metric saw the largest increase, rising by 3.4\%, while the EX metric improved by 2.8\%. This is because the baseline EM metric results contained false negatives (i.e., the generated SQL was correct, but did not match the label SQL literally). Similar to CodeLlama-13b, on the CodeLlama-7b and Wizardcoder-15B models, the improvement in the EM metric was better than that of the EX metric. For CodeLlama-7b, the EX and EM metrics increased by 2.1\% and 2.8\% respectively; on the WizardCoder-15B model, the EX and EM metrics improved by 1.2\% and 2.4\% respectively. It is evident that in four code models, three showed better improvement in the EM metric than the EX metric, which also suggests that the methods proposed in this paper are more conducive to the model generating correct column names and table names, thus reducing mismatches. Moreover, a comparison of the four code models reveals that the choice of base model has a significant impact on the capability of SQL generation; the evaluation results of WizardCoder-15b were only on par with those of CodeLlama-7b, while the performance of deepseek-coder-6.7b even exceeded that of CodeLlama-13b, which also tells us about the importance of choosing the right base model in the training of large domain models. Additionally, the method proposed in this study is not only effective on large code models but also shows an increase of 2.8\% in the EX metric and 2.5\% in the EM metric on the Baichuan2-7B-Chat general model.

Lastly, while many studies mainly report improvements in the EX metric, a decline in the EM metric may occur. This decline can be due to the exactness of execution accuracy only, implying that the query results are the same, but the SQL statements are logically incorrect. However, the method introduced in this paper achieves enhancements in both the EX and EM metrics, with the most notable improvement observed in the EM metric. This underscores that the enhancement in the EX metric primarily results from reducing errors in column names, values, etc., thereby improving the EM metric and ensuring the generated SQL is more accurate.
\begin{table}[htbp]
\caption{\label{citation-guide}
Example of \textsc{Spider-Syn} and \textsc{Spider}
}
\vspace{10pt}
\centering
\resizebox{1\columnwidth}{!}{\begin{tabular}{l}
\toprule
\colorbox{yellow}{\textbf{Spider:}} \texttt{How many \colorbox{green}{singers} do we have?}\\
\colorbox{yellow}{\textbf{Spider-Syn:}}\texttt{How many \colorbox{green}{vocalists} do we have?}\\
\bottomrule
\end{tabular}}
\end{table}

\textbf{Robustness Tests.} We modeled the relationships between column names and cell values as well as those between table names and cell values. The aim was to enhance the model's semantic understanding of column and table names, thereby enabling it to recognize synonyms for these names as expressed in questions. To verify this, we conducted robustness tests using the \textsc{Spider-Syn} dataset\citep{gan2021towards}, which consists of questions from the \textsc{spider-dev} dataset that have been modified with synonym substitutions, resulting in column and table names not appearing directly in the questions. This approach, as demonstrated in Table 3, aligns with the diversity of user questions in real-world scenarios.

We used the same model to predict SQL on the validation set of \textsc{Spider-Syn}, and the results are presented in Table 2. From the results in Table 2, it can be seen that the semantically enhanced model also shows some improvement in robustness, with both \textsc{EX} and \textsc{EM} metrics being higher than the baseline model. This indicates that the semantically enhanced model has a more effective understanding of the semantic information of column and table names.

\textbf{Case Study.}
We primarily focus on case analyses of the results from the CodeLlama-7b-Instruct and Baichuan2-7B-Chat models, as shown in Table 4. The first column presents a case analysis for the Baichuan2-7B-Chat model. From this column, it is evident that the SQL statements generated by the Baichuan2-7B-Chat model without domain database knowledge infusion exhibited value and column errors. For instance, the column mentioned in the question is "\texttt{type of pet}," but the baseline model produced "\texttt{pet\_type}," indicating the model's understanding of the question but unfamiliarity with the database's schema, resulting in an incorrect column name. In contrast, the Baichuan2-7B-Chat model, after domain database knowledge infusion, accurately generated the column name.  \begin{table*}[htbp]
\centering
\caption{\label{citation-guide}Case study on results produced by different models. Discrepancies between the predicted results and the gold are highlighted in green background."\textbf{Predict}" refers to the baseline models without domain database knowledge infusion, generating SQL statements based on the questions and database schema after fine-tuning on the \textsc{Spider} training set. "\textbf{Predict$^{\star}$}" represents the baseline models after domain database knowledge infusion, also generating SQL statements based on the questions and database schema information fine-tuning on the \textsc{Spider} training set. }
\vspace{10pt}
\begin{tabular}{p{15cm}}
\toprule
\textbf{Bad Case of Baichuan2-7B-Chat}\\
\midrule
\multicolumn{1}{c}{\emph{Columns Generation Error}}\\
\emph{\textbf{What type of pet is the youngest animal, and how much does it weigh?}}\\
\colorbox{yellow}{\textbf{Gold:}} \texttt{SELECT \colorbox{pink}{pettype},  weight FROM pets ORDER BY pet\_age LIMIT 1}\\
\colorbox{yellow}{\textbf{Predict:}} \texttt{SELECT \colorbox{green}{pet\_type},  weight FROM pets ORDER BY pet\_age LIMIT 1}\\
\colorbox{yellow}{\textbf{Predict$^{\star}$:}}\texttt{SELECT \colorbox{pink}{pettype}, weight FROM pets ORDER BY pet\_age LIMIT 1}\\
\hdashline
\multicolumn{1}{c}{\emph{Mismatch between Column Name and Cell Value}}\\
\emph{\textbf{Return the number of flights arriving in Aberdeen.}}\\
\colorbox{yellow}{\textbf{Gold:}}\texttt{SELECT count(*) FROM FLIGHTS AS T1 JOIN AIRPORTS AS T2 ON T1.DestAirport = T2.AirportCode WHERE \colorbox{pink}{T2.City} = "Aberdeen"}\\
\colorbox{yellow}{\textbf{Predict:}}\texttt{SELECT count(*) FROM flights WHERE \colorbox{green}{DestAirport} = "Aberdeen"}\\
\colorbox{yellow}{\textbf{Predict$^{\star}$:}}\texttt{SELECT count(*) FROM airports AS T1 JOIN flights AS T2 ON T1.AirportCode = T2.DestAirport WHERE \colorbox{pink}{T1.City} = "Aberdeen"}\\
\toprule
\textbf{Bad Case of CodeLlama-7b-Instruct}\\
\midrule
\multicolumn{1}{c}{\emph{Columns Generation Error}}\\
\emph{\textbf{List the arrival date and the departure date for all the dogs}}\\
\colorbox{yellow}{\textbf{Gold:}} \texttt{SELECT \colorbox{pink}{date\_arrived} ,  \colorbox{pink}{date\_departed} FROM Dogs}\\
\colorbox{yellow}{\textbf{Predict:}} \texttt{SELECT \colorbox{green}{date\_of\_arrival},  \colorbox{green}{date\_of\_departure} FROM Dogs}\\
\colorbox{yellow}{\textbf{Predict$^{\star}$:}}\texttt{SELECT \colorbox{pink}{date\_arrived}, \colorbox{pink}{date\_departed} FROM dogs}\\
\hdashline
\multicolumn{1}{c}{\emph{Mismatch between Column Name and Cell Value}}\\
\emph{\textbf{What country is Jetblue Airways affiliated with?}}\\
\colorbox{yellow}{\textbf{Gold:}}\texttt{SELECT Country FROM AIRLINES WHERE \colorbox{pink}{Airline}  =  "JetBlue Airways"}\\
\colorbox{yellow}{\textbf{Predict:}}\texttt{SELECT Country FROM airlines WHERE \colorbox{green}{Abbreviation}  =  "Jetblue Airways"}\\
\colorbox{yellow}{\textbf{Predict$^{\star}$:}}\texttt{SELECT Country FROM airlines WHERE \colorbox{pink}{Airline}  =  "JetBlue Airways"}\\
\bottomrule
\end{tabular}
\end{table*}Additionally, regarding value errors, the baseline model incorrectly associated the value "\texttt{Aberdeen}," which should belong to the "\texttt{City}" column of the "\texttt{AIRPORTS}" table, with the "\texttt{DestAirport}" column of the "\texttt{flights}" table. This mistake shows the model correctly extracted the value "\texttt{Aberdeen}" from the question but misconnected the column name. Moreover, the logic of this statement did not align with the question, whereas the optimized model correctly generated the statement with logical consistency. This demonstrates that infusing domain database knowledge not only aids in correcting errors related to values and columns but also enhances the model's ability to understand and follow instructions. Similar issues were observed with the CodeLlama-7b-Instruct model.

\begin{table}[htbp]
\centering
\caption{Accuracy results of column name generation,"\emph{KE}" stands for "Knowledge Enhanced," }
\vspace{10pt}
\begin{tabular}{ccc|cc|cc}
\Xhline{1.1pt}
\multirow{2}{*}{Models}   & \multicolumn{2}{c|}{deepseek-coder-6.7b} & \multicolumn{2}{c|}{CodeLlama-7b} & \multicolumn{2}{c}{CodeLlama-13b} \\
\cline{2-7}
         & Baseline             & KE               & Baseline          & KE           & Baseline          & KE            \\
         \hline   
Accuracy & 0.73                 & 0.77             & 0.75              & 0.77         & 0.75              & 0.78     \\ 
\Xhline{1.1pt}
\end{tabular}
\end{table}

\textbf{Evaluating Improvements in the Handling of Hallucination Issues.}
We also performed an error analysis on the column names generated in SQL because we aim to diminish the hallucination problem in LLMs and reduce the errors in generating column and table names by increasing their co-occurrence frequency. For this purpose, we conducted a SQL analysis using deepseek and Codellma as examples. Our calculation method is $\mathcal{C}_{correct}/\mathcal{C}_{total}$, where $\mathcal{C}_{total}$ is the total number of column names generated in the SQL, and $\mathcal{C}_{correct}$ is the number of those generated column names that exist within the schema (meaning the column names are correct and not fabricated by the model).

The results are shown in Table 5. The data in the table reveals that the accuracy of the column names generated by the model, which has been enhanced with domain-specific database knowledge, has improved by 2\% to 4\%. This improvement also suggests a decrease in the generation of incorrect column names resulting from the hallucination problem in LLMs.

\begin{table}[htbp]
\centering
\caption{Performance on Unseen Database}
\vspace{10pt}
\begin{tabular}{lcc}
\toprule
\multirow{2}{*}{\textbf{Models}}                 & \multicolumn{2}{c}{\textsc{Spider-test}}                             \\
\cline{2-3}
\rule{0pt}{12pt}
                                                 & \textbf{EX\%}                    & \textbf{EM\%}                    \\
                                                 \midrule
deepseek-6.7b   & 77.8                           & 73.9           \\
deepseek-6.7b$_{+\emph{KE(ours)}}$  & 79.4                           & 72.1 \\  
CodeLlama-7b                                     & 74.1                           & 67.8                           \\
CodeLlama-7b$_{+\emph{KE(ours)}}$   & 75.5 & 68.2 \\
CodeLlama-13b                                    & 77.7                           & 70.1                           \\
CodeLlama-13b$_{+\emph{KE(ours)}}$   & 76.9 & 70.8 \\
WizardCoder-15B                                  & 73.8                           & 66.0                          \\
WizardCoder-15B$_{+\emph{KE(ours)}}$ & 73.0 & 66.8 \\
\bottomrule
\end{tabular}
\end{table}

\textbf{Performance on Unseen Database.}
During the semantic knowledge injection phase, our model utilized prior knowledge data constructed from the \textsc{Spider} training and development set databases; hence, we did not perform semantic enhancement on the test set database (the databases of the test set and development set do not overlap). We validated our model, which had been subjected to knowledge injection using the training and development set databases, on \textsc{Spider} test set. On one hand, we aimed to verify whether the knowledge-injected model would affect its capability with unknown databases. On the other hand, we wanted to further confirm that the improvement from domain knowledge injection enhances the model's understanding of databases it has learned about, while maintaining its original capabilities with unseen databases. This is because the method of this paper is not aimed at enhancing the model's generalizability, but rather at improving its understanding of domain-specific databases.

The experimental results, as shown in Table 6, indicate that the model's performance on unknown databases has not been impacted and remains similar to that of the baseline model. This also validates the idea presented in this paper.

\begin{table}[htbp]
\centering
\caption{The results of the ablation analysis experiment are as follows: "\emph{schema}" indicates that only the database schema knowledge was infused into the model; "\emph{semantic}" denotes that only semantic knowledge of columns and tables was injected into the model.}
\vspace{10pt}
\begin{tabular}{lcc}

\toprule
\multirow{2}{*}{\textbf{Models}}                 & \multicolumn{2}{c}{\textsc{Spider-dev}}                             \\
\cline{2-3}
\rule{0pt}{12pt}
                                                 & \textbf{EX\%}                    & \textbf{EM\%}                    \\
\hline
CodeLlama-7b   & 74.5 & 70.2 \\
CodeLlama-7b + \emph{schema} & 75.0 & \textbf{71.1} \\
CodeLlama-7b + \emph{semantic} & \textbf{75.1} & 69.9 \\
\bottomrule
\end{tabular}

\label{tab:accents}
\end{table}

\textbf{Ablation Study.}
We conducted an ablation analysis using the CodeLlama-7b-Instruct model as an example, comparing the effects of infusing knowledge related to database values and database schema information. The experimental results are presented in Table 7. The outcomes indicate that executing either task individually offers a certain degree of improvement to the model, yet it does not achieve the optimal situation.

Furthermore, we can also see that after the injection of schema knowledge, the model's EM metric improves more noticeably. This precisely demonstrates that increasing the co-occurrence frequency of table and column names aids the model in generating the correct column and table names.
\begin{table}[htbp]
\centering
\caption{The results of training with a mixture of domain knowledge infusion data and \textsc{Spider} training set data.}
\vspace{10pt}
\begin{tabular}{lcc}
\toprule
\multirow{2}{*}{\textbf{Models}}                 & \multicolumn{2}{c}{\textsc{Spider-dev}}                             \\
\cline{2-3}
\rule{0pt}{12pt}
                                                 & \textbf{EX\%}                    & \textbf{EM\%}                    \\
\hline
CodeLlama-7b$_{+\emph{KE(ours)}}$               & \textbf{76.6} & \textbf{73.0} \\
CodeLlama-7b$_{+mixed}$ & 73.8 & 68.5 \\
Baichuan2-7B$_{+\emph{KE(ours)}}$               & \textbf{71.3} & \textbf{66.5}   \\
Baichuan2-7B$_{+mixed}$ & 67.0 & 61.9 \\
\hline
\end{tabular}

\label{tab:accents}
\end{table}

\textbf{Necessity of Phased Training.}
We mixed domain knowledge data with training data for the Text-to-SQL tasks and trained them together. When comparing the effects of integrating the knowledge injection task with the Text-to-SQL task, we found that the combination affected the model's performance. As shown in table 8. This decline can be attributed to the misalignment between the training objectives of upstream database knowledge infusion tasks and downstream Text-to-SQL tasks, coupled with the predominance of upstream task data preventing sufficient training for the downstream task. This also demonstrates the necessity of phased training.

\section{Related Work}
In Text-to-SQL tasks, controlling the correctness of generated column names and values due to the model's unpredictability has always been a challenge. Numerous studies focus on ensuring the accuracy of column names, value joins, and syntactic correctness. These efforts differ from the approach proposed in this paper; they either control at the decoding end, such as through syntax trees \citep{scholak2021picard,guo2019towards,yin-neubig-2017-syntactic}, or incorporate schema linking information during the fine-tuning phase of downstream tasks \citep{wang2019rat}. Additionally, some utilize graph neural networks to connect the relationships between data tables and columns \citep{bogin2019representing,cao2021lgesql,huang2021relation}. Our method focuses on injecting knowledge into the base model to address these issues, thereby also enhancing the base model's understanding of databases.

Many studies improve the model's table comprehension abilities through pre-training on massive tabular data \citep{yin2020tabert,yu2020grappa,deng2020structure}. These methods require significant computational resources for pre-training on vast tabular data and lack specificity. In contrast, our approach specifically injects knowledge into domain databases, making it more targeted and resource-efficient.

The volume of Text-to-SQL work related to Large Language Models (LLMs) is steadily increasing, with many studies achieving impressive results through in-context learning \citep{brown2020language,pourreza2023din,liu2023comprehensive} or Chain-of-thought approaches \citep{wei2022chain,pourreza2023din} on platforms like ChatGPT. Furthermore, there are efforts focused on fine-tuning pre-trained models \citep{shaw2020compositional,xie2022unifiedskg,qi2022rasat}, with current state-of-the-art (SOTA) methods based on fine-tuning achieving notable success \citep{li2023resdsql,rai2023improving}, utilizing models with up to 3 billion parameters. However, methods involving even larger scale models require further exploration. Our proposed method primarily experiments on larger scale open-source models, providing a new experimental baseline for open-source LLMs.

\section{Conclusion and Future Work}
We propose a domain database knowledge injection method that incorporates the semantic information of column names, table names, and schema into LLMs through the design of three training objectives. This method successfully elevates the EX and EM scores, but it does come with certain challenges and constraints. Firstly, training with data from databases may pose privacy risks if it involves users' sensitive data, potentially leading to privacy breaches. Secondly, we observed that even with knowledge injection, the model's comprehension of complex instructions and the quality of generated complex SQL queries are still not satisfactory. This suggests that our method has limited effectiveness in improving instruction understanding and addressing complex problems, necessitating the optimization of instruction comprehension capabilities through additional approaches. Lastly, the proposed method in this paper, due to the addition of an extra training stage, also incurs increased training and time costs.

In future work, we aim to utilize privacy protection techniques to safeguard sensitive data\citep{plant2021cape,shi2021selective,anil2021large}. Additionally, we plan to explore methods such as CoT (Chain-of-Thought)\citep{wei2022chain,zhang2022automatic} to optimize instruction comprehension and reasoning abilities. We will also consider how to inject knowledge during the downstream task training stage, thereby reducing training costs.

\begin{ack}
This work was supported by National Science and Technology Major Project(Grant No.2021ZD0114600), National Key R\&D Program of China (Grant No. 2021YFB2012202), the Hubei Provincial Science
and Technology Major Project of China (Grant No. 2020AEA011), the Key Research \& Development Plan of Hubei Province of
China (Grant No. 2021BAA038), the project of Science,Technology and Innovation Commission of Shenzhen Municipality of China
(Grant No. JCYJ20210324120002006 and JSGG20210802153009028) and National Natural Science Foundation of China (Grant No. 62276260)
\end{ack}





\bibliography{mybibfile}

\begin{thebibliography}{34}
\providecommand{\natexlab}[1]{#1}
\providecommand{\url}[1]{\texttt{#1}}
\expandafter\ifx\csname urlstyle\endcsname\relax
  \providecommand{\doi}[1]{doi: #1}\else
  \providecommand{\doi}{doi: \begingroup \urlstyle{rm}\Url}\fi

\bibitem[Anil et~al.(2021)Anil, Ghazi, Gupta, Kumar, and Manurangsi]{anil2021large}
R.~Anil, B.~Ghazi, V.~Gupta, R.~Kumar, and P.~Manurangsi.
\newblock Large-scale differentially private bert.
\newblock \emph{arXiv preprint arXiv:2108.01624}, 2021.

\bibitem[Bogin et~al.(2019)Bogin, Gardner, and Berant]{bogin2019representing}
B.~Bogin, M.~Gardner, and J.~Berant.
\newblock Representing schema structure with graph neural networks for text-to-sql parsing.
\newblock \emph{arXiv preprint arXiv:1905.06241}, 2019.

\bibitem[Brown et~al.(2020)Brown, Mann, Ryder, Subbiah, Kaplan, Dhariwal, Neelakantan, Shyam, Sastry, Askell, et~al.]{brown2020language}
T.~Brown, B.~Mann, N.~Ryder, M.~Subbiah, J.~D. Kaplan, P.~Dhariwal, A.~Neelakantan, P.~Shyam, G.~Sastry, A.~Askell, et~al.
\newblock Language models are few-shot learners.
\newblock \emph{Advances in neural information processing systems}, 33:\penalty0 1877--1901, 2020.

\bibitem[Cao et~al.(2021)Cao, Chen, Chen, Zhao, Zhu, and Yu]{cao2021lgesql}
R.~Cao, L.~Chen, Z.~Chen, Y.~Zhao, S.~Zhu, and K.~Yu.
\newblock Lgesql: line graph enhanced text-to-sql model with mixed local and non-local relations.
\newblock \emph{arXiv preprint arXiv:2106.01093}, 2021.

\bibitem[Deng et~al.(2020)Deng, Awadallah, Meek, Polozov, Sun, and Richardson]{deng2020structure}
X.~Deng, A.~H. Awadallah, C.~Meek, O.~Polozov, H.~Sun, and M.~Richardson.
\newblock Structure-grounded pretraining for text-to-sql.
\newblock \emph{arXiv preprint arXiv:2010.12773}, 2020.

\bibitem[Gan et~al.(2021)Gan, Chen, Huang, Purver, Woodward, Xie, and Huang]{gan2021towards}
Y.~Gan, X.~Chen, Q.~Huang, M.~Purver, J.~R. Woodward, J.~Xie, and P.~Huang.
\newblock Towards robustness of text-to-sql models against synonym substitution.
\newblock \emph{arXiv preprint arXiv:2106.01065}, 2021.

\bibitem[Guo et~al.(2024)Guo, Zhu, Yang, Xie, Dong, Zhang, Chen, Bi, Wu, Li, et~al.]{guo2024deepseek}
D.~Guo, Q.~Zhu, D.~Yang, Z.~Xie, K.~Dong, W.~Zhang, G.~Chen, X.~Bi, Y.~Wu, Y.~Li, et~al.
\newblock Deepseek-coder: When the large language model meets programming--the rise of code intelligence.
\newblock \emph{arXiv preprint arXiv:2401.14196}, 2024.

\bibitem[Guo et~al.(2019)Guo, Zhan, Gao, Xiao, Lou, Liu, and Zhang]{guo2019towards}
J.~Guo, Z.~Zhan, Y.~Gao, Y.~Xiao, J.-G. Lou, T.~Liu, and D.~Zhang.
\newblock Towards complex text-to-sql in cross-domain database with intermediate representation.
\newblock \emph{arXiv preprint arXiv:1905.08205}, 2019.

\bibitem[Huang et~al.(2021)Huang, Wang, Wang, Dong, and Xiao]{huang2021relation}
J.~Huang, Y.~Wang, Y.~Wang, Y.~Dong, and Y.~Xiao.
\newblock Relation aware semi-autoregressive semantic parsing for nl2sql.
\newblock \emph{arXiv preprint arXiv:2108.00804}, 2021.

\bibitem[Kang and Choi(2023)]{kang2023impact}
C.~Kang and J.~Choi.
\newblock Impact of co-occurrence on factual knowledge of large language models.
\newblock \emph{arXiv preprint arXiv:2310.08256}, 2023.

\bibitem[Li et~al.(2023)Li, Zhang, Li, and Chen]{li2023resdsql}
H.~Li, J.~Zhang, C.~Li, and H.~Chen.
\newblock Resdsql: Decoupling schema linking and skeleton parsing for text-to-sql.
\newblock In \emph{Proceedings of the AAAI Conference on Artificial Intelligence}, volume~37, pages 13067--13075, 2023.

\bibitem[Liu et~al.(2023)Liu, Hu, Wen, and Yu]{liu2023comprehensive}
A.~Liu, X.~Hu, L.~Wen, and P.~S. Yu.
\newblock A comprehensive evaluation of chatgpt's zero-shot text-to-sql capability.
\newblock \emph{arXiv preprint arXiv:2303.13547}, 2023.

\bibitem[Luo et~al.(2023)Luo, Xu, Zhao, Sun, Geng, Hu, Tao, Ma, Lin, and Jiang]{luo2023wizardcoder}
Z.~Luo, C.~Xu, P.~Zhao, Q.~Sun, X.~Geng, W.~Hu, C.~Tao, J.~Ma, Q.~Lin, and D.~Jiang.
\newblock Wizardcoder: Empowering code large language models with evol-instruct.
\newblock \emph{arXiv preprint arXiv:2306.08568}, 2023.

\bibitem[Lyu et~al.(2021)Lyu, Chen, Zhu, and Yu]{lyu2021let}
B.~Lyu, L.~Chen, S.~Zhu, and K.~Yu.
\newblock Let: Linguistic knowledge enhanced graph transformer for chinese short text matching.
\newblock In \emph{Proceedings of the AAAI Conference on Artificial Intelligence}, volume~35, pages 13498--13506, 2021.

\bibitem[Plant et~al.(2021)Plant, Gkatzia, and Giuffrida]{plant2021cape}
R.~Plant, D.~Gkatzia, and V.~Giuffrida.
\newblock Cape: Context-aware private embeddings for private language learning.
\newblock \emph{arXiv preprint arXiv:2108.12318}, 2021.

\bibitem[Pourreza and Rafiei(2023)]{pourreza2023din}
M.~Pourreza and D.~Rafiei.
\newblock Din-sql: Decomposed in-context learning of text-to-sql with self-correction.
\newblock \emph{arXiv preprint arXiv:2304.11015}, 2023.

\bibitem[Qi et~al.(2022)Qi, Tang, He, Wan, Cheng, Zhou, Wang, Zhang, and Lin]{qi2022rasat}
J.~Qi, J.~Tang, Z.~He, X.~Wan, Y.~Cheng, C.~Zhou, X.~Wang, Q.~Zhang, and Z.~Lin.
\newblock Rasat: Integrating relational structures into pretrained seq2seq model for text-to-sql.
\newblock \emph{arXiv preprint arXiv:2205.06983}, 2022.

\bibitem[Raffel et~al.(2020)Raffel, Shazeer, Roberts, Lee, Narang, Matena, Zhou, Li, and Liu]{raffel2020exploring}
C.~Raffel, N.~Shazeer, A.~Roberts, K.~Lee, S.~Narang, M.~Matena, Y.~Zhou, W.~Li, and P.~J. Liu.
\newblock Exploring the limits of transfer learning with a unified text-to-text transformer.
\newblock \emph{Journal of machine learning research}, 21\penalty0 (140):\penalty0 1--67, 2020.

\bibitem[Rai et~al.(2023)Rai, Wang, Zhou, and Yao]{rai2023improving}
D.~Rai, B.~Wang, Y.~Zhou, and Z.~Yao.
\newblock Improving generalization in language model-based text-to-sql semantic parsing: Two simple semantic boundary-based techniques.
\newblock \emph{arXiv preprint arXiv:2305.17378}, 2023.

\bibitem[Roziere et~al.(2023)Roziere, Gehring, Gloeckle, Sootla, Gat, Tan, Adi, Liu, Remez, Rapin, et~al.]{roziere2023code}
B.~Roziere, J.~Gehring, F.~Gloeckle, S.~Sootla, I.~Gat, X.~E. Tan, Y.~Adi, J.~Liu, T.~Remez, J.~Rapin, et~al.
\newblock Code llama: Open foundation models for code.
\newblock \emph{arXiv preprint arXiv:2308.12950}, 2023.

\bibitem[Scholak et~al.(2021)Scholak, Schucher, and Bahdanau]{scholak2021picard}
T.~Scholak, N.~Schucher, and D.~Bahdanau.
\newblock Picard: Parsing incrementally for constrained auto-regressive decoding from language models.
\newblock \emph{arXiv preprint arXiv:2109.05093}, 2021.

\bibitem[Shaw et~al.(2020)Shaw, Chang, Pasupat, and Toutanova]{shaw2020compositional}
P.~Shaw, M.-W. Chang, P.~Pasupat, and K.~Toutanova.
\newblock Compositional generalization and natural language variation: Can a semantic parsing approach handle both?
\newblock \emph{arXiv preprint arXiv:2010.12725}, 2020.

\bibitem[Shi et~al.(2021{\natexlab{a}})Shi, Ng, Wang, Zhu, Li, Wang, dos Santos, and Xiang]{shi2021learning}
P.~Shi, P.~Ng, Z.~Wang, H.~Zhu, A.~H. Li, J.~Wang, C.~N. dos Santos, and B.~Xiang.
\newblock Learning contextual representations for semantic parsing with generation-augmented pre-training.
\newblock In \emph{Proceedings of the AAAI Conference on Artificial Intelligence}, volume~35, pages 13806--13814, 2021{\natexlab{a}}.

\bibitem[Shi et~al.(2021{\natexlab{b}})Shi, Cui, Li, Jia, and Yu]{shi2021selective}
W.~Shi, A.~Cui, E.~Li, R.~Jia, and Z.~Yu.
\newblock Selective differential privacy for language modeling.
\newblock \emph{arXiv preprint arXiv:2108.12944}, 2021{\natexlab{b}}.

\bibitem[Wang et~al.(2019)Wang, Shin, Liu, Polozov, and Richardson]{wang2019rat}
B.~Wang, R.~Shin, X.~Liu, O.~Polozov, and M.~Richardson.
\newblock Rat-sql: Relation-aware schema encoding and linking for text-to-sql parsers.
\newblock \emph{arXiv preprint arXiv:1911.04942}, 2019.

\bibitem[Wang et~al.(2022)Wang, Kordi, Mishra, Liu, Smith, Khashabi, and Hajishirzi]{wang2022self}
Y.~Wang, Y.~Kordi, S.~Mishra, A.~Liu, N.~A. Smith, D.~Khashabi, and H.~Hajishirzi.
\newblock Self-instruct: Aligning language models with self-generated instructions.
\newblock \emph{arXiv preprint arXiv:2212.10560}, 2022.

\bibitem[Wei et~al.(2022)Wei, Wang, Schuurmans, Bosma, Xia, Chi, Le, Zhou, et~al.]{wei2022chain}
J.~Wei, X.~Wang, D.~Schuurmans, M.~Bosma, F.~Xia, E.~Chi, Q.~V. Le, D.~Zhou, et~al.
\newblock Chain-of-thought prompting elicits reasoning in large language models.
\newblock \emph{Advances in Neural Information Processing Systems}, 35:\penalty0 24824--24837, 2022.

\bibitem[Xie et~al.(2022)Xie, Wu, Shi, Zhong, Scholak, Yasunaga, Wu, Zhong, Yin, Wang, et~al.]{xie2022unifiedskg}
T.~Xie, C.~H. Wu, P.~Shi, R.~Zhong, T.~Scholak, M.~Yasunaga, C.-S. Wu, M.~Zhong, P.~Yin, S.~I. Wang, et~al.
\newblock Unifiedskg: Unifying and multi-tasking structured knowledge grounding with text-to-text language models.
\newblock \emph{arXiv preprint arXiv:2201.05966}, 2022.

\bibitem[Yang et~al.(2023)Yang, Xiao, Wang, Zhang, Bian, Yin, Lv, Pan, Wang, Yan, et~al.]{yang2023baichuan}
A.~Yang, B.~Xiao, B.~Wang, B.~Zhang, C.~Bian, C.~Yin, C.~Lv, D.~Pan, D.~Wang, D.~Yan, et~al.
\newblock Baichuan 2: Open large-scale language models.
\newblock \emph{arXiv preprint arXiv:2309.10305}, 2023.

\bibitem[Yin and Neubig(2017)]{yin-neubig-2017-syntactic}
P.~Yin and G.~Neubig.
\newblock A syntactic neural model for general-purpose code generation.
\newblock In R.~Barzilay and M.-Y. Kan, editors, \emph{Proceedings of the 55th Annual Meeting of the Association for Computational Linguistics (Volume 1: Long Papers)}, pages 440--450, Vancouver, Canada, July 2017. Association for Computational Linguistics.
\newblock \doi{10.18653/v1/P17-1041}.
\newblock URL \url{https://aclanthology.org/P17-1041}.

\bibitem[Yin et~al.(2020)Yin, Neubig, Yih, and Riedel]{yin2020tabert}
P.~Yin, G.~Neubig, W.-t. Yih, and S.~Riedel.
\newblock Tabert: Pretraining for joint understanding of textual and tabular data.
\newblock \emph{arXiv preprint arXiv:2005.08314}, 2020.

\bibitem[Yu et~al.(2018)Yu, Zhang, Yang, Yasunaga, Wang, Li, Ma, Li, Yao, Roman, et~al.]{yu2018spider}
T.~Yu, R.~Zhang, K.~Yang, M.~Yasunaga, D.~Wang, Z.~Li, J.~Ma, I.~Li, Q.~Yao, S.~Roman, et~al.
\newblock Spider: A large-scale human-labeled dataset for complex and cross-domain semantic parsing and text-to-sql task.
\newblock \emph{arXiv preprint arXiv:1809.08887}, 2018.

\bibitem[Yu et~al.(2020)Yu, Wu, Lin, Wang, Tan, Yang, Radev, Socher, and Xiong]{yu2020grappa}
T.~Yu, C.-S. Wu, X.~V. Lin, B.~Wang, Y.~C. Tan, X.~Yang, D.~Radev, R.~Socher, and C.~Xiong.
\newblock Grappa: Grammar-augmented pre-training for table semantic parsing.
\newblock \emph{arXiv preprint arXiv:2009.13845}, 2020.

\bibitem[Zhang et~al.(2022)Zhang, Zhang, Li, and Smola]{zhang2022automatic}
Z.~Zhang, A.~Zhang, M.~Li, and A.~Smola.
\newblock Automatic chain of thought prompting in large language models.
\newblock \emph{arXiv preprint arXiv:2210.03493}, 2022.

\end{thebibliography}

\end{document}